\documentclass[letterpaper, 10 pt, conference]{Settings/ieeeconf}  

\IEEEoverridecommandlockouts                              

\usepackage{graphics} 
\usepackage{epsfig} 
\usepackage{times} 
\usepackage{amsmath} 
\usepackage{amssymb}  
\usepackage{svg}
\usepackage{comment}
\usepackage{url}

\usepackage{breakurl}
\usepackage{graphicx}
\usepackage{subcaption}
\usepackage{tabularx}
\usepackage{array} 
\usepackage{multirow} 
\usepackage{booktabs}
\usepackage{siunitx}
\usepackage{hyperref}

\title{\LARGE \bf
ARMimic: Learning Robotic Manipulation from Passive Human Demonstrations in Augmented Reality
}

\author{Rohan Walia$^{1}$, Yusheng Wang$^{2}$, Ralf Römer$^{1}$, \\ Masahiro Nishio$^{3}$, Angela P. Schoellig$^{1}$, Jun Ota$^{2}$%
\thanks{$^{1}$ Department of Computer Engineering; \href{https://www.dynsyslab.org/}{Learning Systems and Robotics Lab}; Munich Institute of Robotics and Machine Intelligence (MIRMI), Technical University of Munich, Germany}%
\thanks{$^{2}$ Department of Precision Engineering; \href{https://otalab.race.t.u-tokyo.ac.jp/en/}{Mobile Robotics Lab}; Research into Artifacts, Center for Engineering (RACE), University of Tokyo, Japan}%
\thanks{$^{3}$ Toyota Motor Corporation, Japan}%
\thanks{\hspace*{-0.4em} Email: {\tt rohan.walia@tum.de}, {\tt wang@robot.t.u-tokyo.ac.jp}}
}

\begin{document}

\maketitle
\thispagestyle{empty}
\pagestyle{empty}


\

\begin{abstract}

Imitation learning is a powerful paradigm for robot skill acquisition, yet conventional demonstration methods—such as kinesthetic teaching and teleoperation—are cumbersome, hardware-heavy, and disruptive to workflows. 
Recently, passive observation using extended reality (XR) headsets has shown promise for egocentric demonstration collection, yet current approaches require additional hardware, complex calibration, or constrained recording conditions that limit scalability and usability.
We present ARMimic, a novel framework that overcomes these limitations with a lightweight and hardware-minimal setup for scalable, robot-free data collection using only a consumer XR headset and a stationary workplace camera.
ARMimic integrates egocentric hand tracking, augmented reality~(AR) robot overlays, and real-time depth sensing to ensure collision-aware, kinematically feasible demonstrations. A unified imitation learning pipeline is at the core of our method, treating both human and virtual robot trajectories as interchangeable, which enables policies that generalize across different embodiments and environments.
We validate ARMimic on two manipulation tasks, including challenging long-horizon bowl stacking. 
In our experiments, ARMimic reduces demonstration time by~50\% compared to teleoperation and improves task success by 11\% over ACT, a state-of-the-art baseline trained on teleoperated data.
Our results demonstrate that ARMimic enables safe, seamless, and in-the-wild data collection, offering great potential for scalable robot learning in diverse real-world settings.
\end{abstract}

\section{INTRODUCTION}

Robot learning for manipulation is central to the development of autonomous systems, enabling their deployment in increasingly dynamic and diverse environments. 
With the growing accessibility of robotic hardware and the rapid advancement of machine learning techniques, research has made significant strides in tackling the core challenge of imparting complex sensorimotor skills to robots~\cite{chi2023diffusion, zhao_learning_2023, barreiros2025careful}.

A prominent approach to this problem is Imitation Learning---also known as Learning from Demonstration (LfD)---in which robots learn new skills by observing expert demonstrations and inferring a policy that maps observations to actions~\cite{hua_learning_2021}. Demonstrations can be collected using three primary methods~\cite{ravichandar_recent_2020}:

\begin{enumerate}
    \item Teleoperation, where the human directly controls the robot via a joystick, GUI, haptic interface, or virtual reality system;
    \item Kinesthetic teaching, where the human physically guides the robot through the motion while the robot records its state via onboard sensors;
    \item Passive observation, where the human performs the task using their own body, often captured through wearable visual or mechanical tracking devices.
\end{enumerate}

Among these methods, teleoperation has gained popularity~\cite{lin_learning_2024, qin_anyteleop:_2023} due to its direct interaction with the robot, which does not require mapping human demonstrations to the robot’s state or action space---a challenge inherent in passive observation. Teleoperation is generally more convenient compared to the physical effort required in kinesthetic teaching.
Although recent open-source platforms~\cite{wu_gello:_2023, zhao_learning_2023, aloha_2_team_aloha_2024} have made teleoperation more accessible, setup remains time-consuming and often disrupts real-world workflows.

In contrast, passive observation offers greater flexibility, as it is largely software-driven and does not require physical access to the robot during data collection. It is also generally more intuitive for humans to perform a task directly themselves rather than indirectly through a robot, which reduces the time required and makes the process more accessible to novices. 
However, passive observation presents unique challenges, particularly regarding safety measures like collision avoidance and robust mapping to robotic embodiments---especially for dexterous manipulation~\cite{orbik_human_2021, qin_dexmv:_2021}. Recent efforts have sought to leverage passive human video data to extract high-level intent for planning, while relying on separately trained low-level controllers~\cite{wang_mimicplay:_2023, bharadhwaj_towards_2023}. However, the overall performance is ultimately limited by the quality of the low-level policies, which are usually trained on teleoperated data.
The emergence of Extended Reality (XR) headsets with egocentric video and integrated tracking capabilities introduces a new paradigm for passive data collection.
As demonstrated in multiple recent works~\cite{chen_arcap:_2024, liu2025egozero, jiang2025iris}, these devices offer the potential to gather large-scale, high-fidelity demonstrations in natural settings without disrupting real-world workflows, paving the way for more scalable and robust robot learning pipelines.

These approaches demonstrate that combining human motion data with related robot information through egocentric collection can help novice users produce robot-executable demonstrations, support learning across different embodiments, and improve generalization to new environments. However, current frameworks still face limitations: Many require additional hardware such as motion-capture gloves~\cite{chen_arcap:_2024}, involve calibration-intensive setups~\cite{jiang2025iris}, or depend on constrained recording conditions~\cite{liu2025egozero} that reduce adaptability.
In real-world settings, workspace constraints and physical barriers further complicate trajectory feasibility and collision assessment, making reliable passive observation and generalization an open research problem.

To address these limitations, we introduce ARMimic, a novel framework for safe, hardware-minimalistic, and scalable data collection using an XR headset like the Meta Quest 3~(Meta Platforms, Inc., Menlo Park, CA, USA)~\cite{meta_quest3}. Our core contribution lies in the unique combination of the following features:
\begin{itemize}
    \item \textbf{Lightweight and portable setup:} 
    We leverage the XR headset's native hand tracking to capture the user's hand pose, which is used to infer the robot's end effector pose. This eliminates the need for cumbersome motion-capture gloves or complex external camera arrays. To provide a complete, robot-agnostic view of the workspace, our setup only requires a single stationary camera.
    \item \textbf{Virtual embodiment and data alignment:} ARMimic reconstructs and visualizes the full robot configuration in real time as a virtual overlay within the headset, which is kinematically aligned with the user's hand movements. 
    We align human and robot data by masking both the human hand and the virtual robot in the visual stream, creating a unified representation that is agnostic to whether the demonstration is human or robot-centric and enabling policy learning from human and robot data.
    \item \textbf{Real-time collision handling and embodiment correction:} 
    ARMimic performs real-time collision detection for the entire virtual robot without requiring pre-scanned environments.
    When a collision or unreachable configuration is detected, the user can switch between inverse kinematic solutions that preserve the end effector pose.
    This mechanism not only ensures safe interaction with the environment but also improves alignment between human and robot embodiments, enhancing data quality and robustness.
\end{itemize}
ARMimic 
enables non-expert users to collect high-quality demonstrations during routine tasks using only an XR headset combined with a workspace camera. By leveraging native tracking capabilities and removing the need for physical robots or external motion-capture systems, our framework minimizes disruption to workflows and seamlessly integrates into existing environments.
We validate ARMimic on common manipulation tasks, demonstrating faster and more intuitive data collection that results in improved task speed and performance compared to learning based on teleoperation.


\section{RELATED WORK}

Recent large-scale cross-embodiment datasets~\cite{o2024open, jiang_galaxea_2025, walke_bridgedata_2023} and models~\cite{shukor2025smolvla, black2024pi_0, doshi_scaling_2024} underscore the benefits of unifying demonstrations across different robots and environments. At the same time, the high cost of collecting such datasets motivates leveraging robot-free data sources that scale more naturally.

Early efforts in this area focused on learning from passive human videos, where agents observe humans interacting with objects. For instance, DexMV~\cite{qin_dexmv:_2021} used hand keypoints extracted from monocular human videos to enable imitation of dexterous manipulation skills, an approach adopted by other works as well~\cite{haldar2025point, Li2022MetaImitationLB, machines10111049}. 
MimicPlay~\cite{wang_mimicplay:_2023} infers task intent from play-based demonstrations, while frameworks such as Towards Human-to-Robot Video Imitation~\cite{bharadhwaj_towards_2023} extract high-level affordances from human videos. 
Some approaches attempt to bridge the domain gap between humans and robots using specialized gripper-shaped gloves~\cite{chi_universal_2024, kim_training_2023}. More recently, NIL~\cite{albaba2025nil} adapted pretrained video diffusion models for imitation learning without robot-specific demonstrations, but embodiment transfer and safe deployment remain challenging.

The advent of XR headsets has enabled a new paradigm for scalable egocentric data collection. Frameworks like ARCap~\cite{chen_arcap:_2024}, RAMPA~\cite{dogangun_rampa:_2025}, and dARt Vinci~\cite{liu_dart_2025}, guide users with real-time augmented reality~(AR) robot overlays and provide warnings for kinematic inconsistencies and potential collisions. 
Other recent works~\cite{kareer_egomimic:_2024, liu2025immimic, niu_human2locoman:_2025, qiu_humanoid_2025}, such as EgoMimic~\cite{kareer_egomimic:_2024}, co-train policies on both egocentric human hand motion and teleoperated robot data to improve generalization across embodiments. Extensions of this paradigm include EgoZero~\cite{liu2025egozero}, which demonstrates large-scale data collection with smart glasses in natural environments, and IRIS~\cite{jiang2025iris}, which provides immersive AR interfaces for robot instruction.

Despite their impressive capabilities, these frameworks suffer from complex calibration, limited portability, and insufficient safety measures, constraining robust dataset collection. Thus, developing a hardware-minimalistic approach that enables safe, large-scale deployment in real-world environments remains an open challenge, and ARMimic specifically aims at addressing these problems.

\section{METHODS}\label{sec:methods}

This section presents a detailed description of ARMimic, starting with the design and implementation of the data collection setup, followed by the learning architecture as well as the preprocessing steps to achieve cross-domain alignment, and concluding with deployment considerations.

\begin{figure*}[tbh!]
\vspace{1.7mm}
  \centering
  \includegraphics[width=0.8\textwidth]{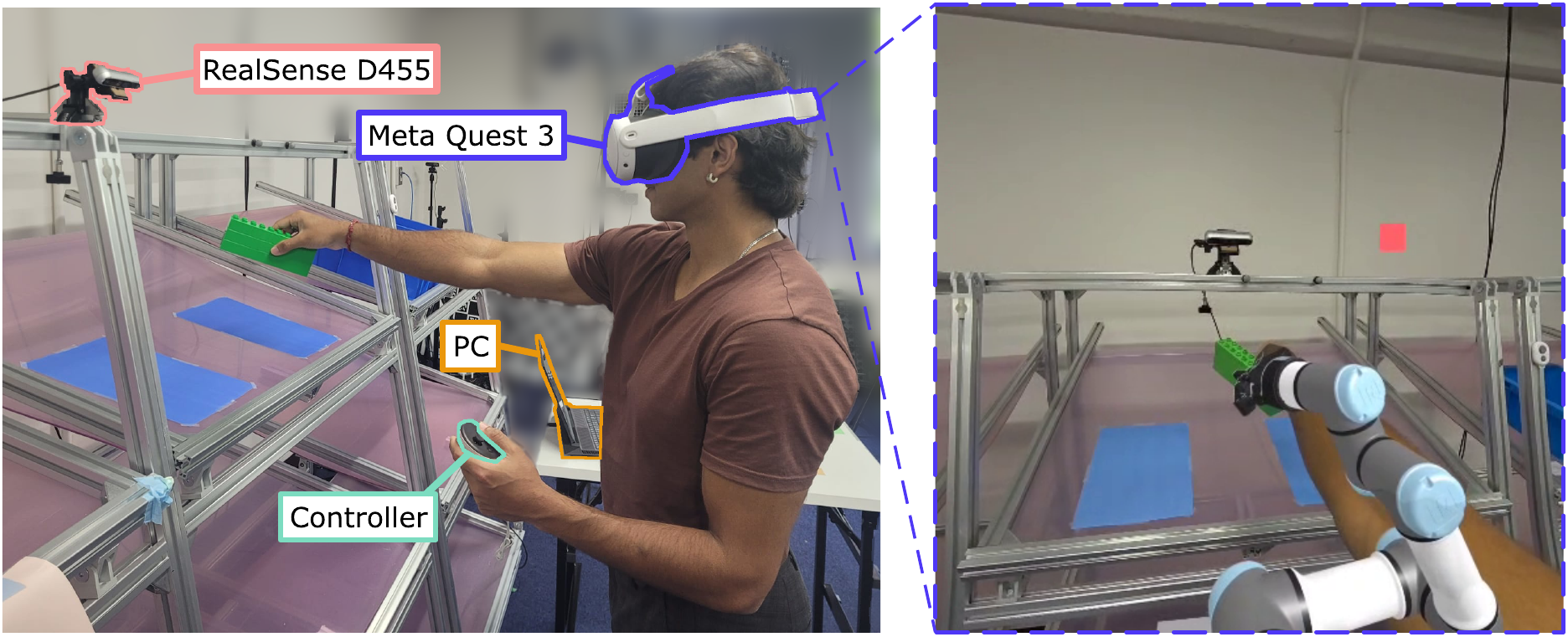}
  \caption{Data collection system with headset view, where the red mark indicates an active recording.}
  \label{figure_system}
\end{figure*}

\subsection{Data Collection}

\subsubsection{System Overview}  
The data collection setup (see Figure~\ref{figure_system}) comprises an XR headset and a stationary RGB-D camera mounted to overlook the workspace. ARMimic leverages the device's native hand tracking to estimate the 6-DoF pose~$p_t$ of the user's right hand in real time. 
We use the pose estimate to compute the full joint configuration~$q_t$ of a virtual robot via inverse kinematics, enabling a semi-transparent AR robot overlay that aligns with the user’s hand motion. 
We consider robots equipped with a parallel gripper and perform gripper control for the AR robot overlay via the distance between the user's right-hand thumb and index fingertip, allowing for intuitive and responsive manipulation within the augmented environment.

During demonstration recording, the XR headset streams the headset’s camera pose, the user's right-hand pose, proprioceptive information from the virtual robot, the passthrough image, i.e., the real-world view without any VR components and a rendered image of the robot overlay.
Simultaneously, we record synchronized visual and depth data from the RGB-D camera. We temporally align all data streams during postprocessing to construct a unified multimodal demonstration dataset.

\subsubsection{User Experience and Feedback} 
\begin{figure}[h!]
  \centering
  \begin{tabular}[t]{p{0.6\linewidth} p{0.37\linewidth}}
    \vspace{0pt} 
    \begin{itemize}
      \item \textbf{Trigger:} Cycles through up to six possible inverse kinematics solutions for the robot joint configuration~$q(t)$ corresponding to the same end effector pose.
      \item \textbf{Grip:} Calibrates the position of the robot base at the beginning of a demonstration set using the joystick.
      \item \textbf{Y button:} Starts recording a demonstration.
      \item \textbf{X button:} Ends the recording.
    \end{itemize}
    &
    \vspace{3mm} 
    \centering
    \includegraphics[width=0.9\linewidth]{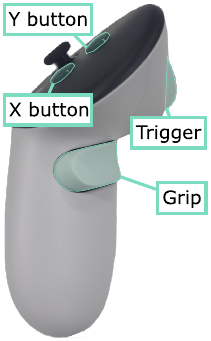}
  \end{tabular}
  \label{figure_controller}
  \caption{Intuitive controller input mapping.}
\label{fig:controller_mapping}
\end{figure}

While one of the user's hands must be used for manipulation via pose tracking, the other hand can hold a controller to easily handle the recording and robot configuration, as visualized in Figure~\ref{figure_controller}. 
The mapping of the controller inputs may be defined according to user preferences, but we found the setup shown in Figure~\ref{fig:controller_mapping} to be intuitive to use and work well in practice.

\begin{figure}[tb!]
    \centering
    \begin{subfigure}{\linewidth}
        \centering
        \includegraphics[width=\linewidth]{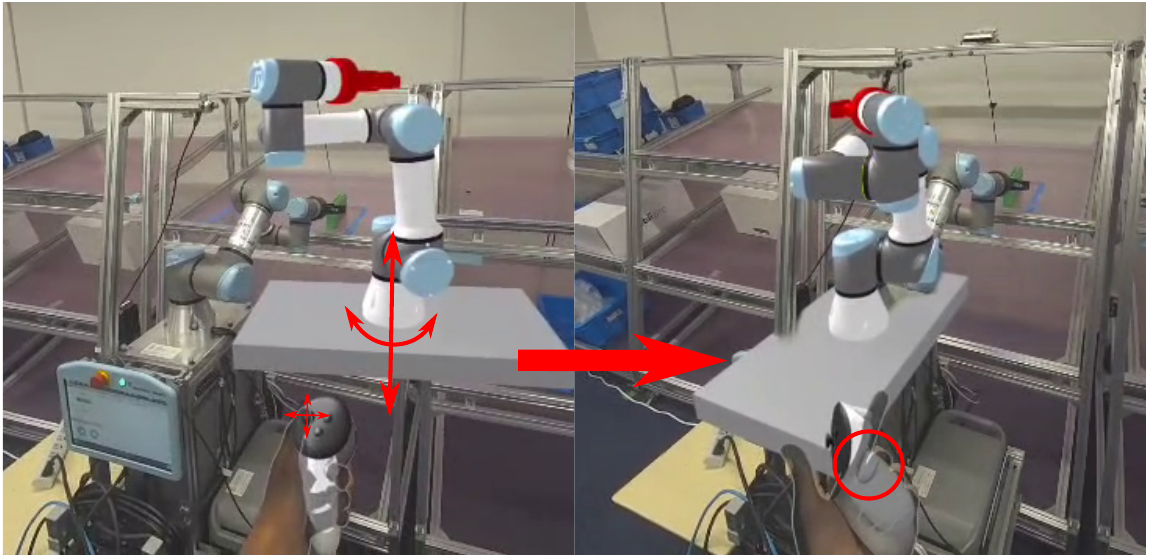}
        \caption{Adjusting the robot base pose via joystick and gripper button.}
        \label{fig:controller_a}
    \end{subfigure}
    
    \vspace{1ex} 
    
    \begin{subfigure}{\linewidth}
        \centering
        \includegraphics[width=\linewidth]{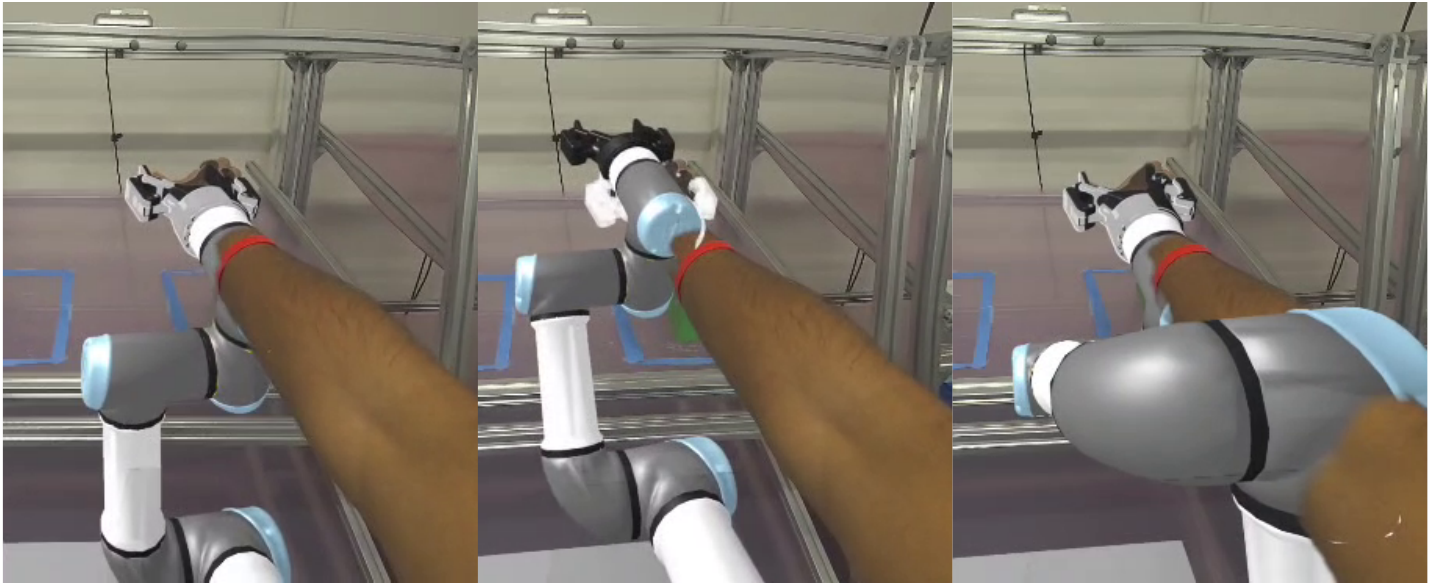}
        \caption{Switching between inverse kinematic solutions via trigger button.}
        \label{fig:controller_b}
    \end{subfigure}
    
    \caption{User input controlling robot configuration.}
    \label{figure_controller}
\end{figure}

The system provides real-time visual feedback within the headset’s field of view, indicating key states such as active demonstration recording or an infeasible robot trajectory. 
We safeguard against these trajectories, which may result from invalid inverse kinematics solutions leading to singularities or self-collisions, by monitoring the kinematics and rigid-body model of the virtual robot.
Collision checks are performed by monitoring the distance between the virtual robot's rigid body model and the environment's point cloud derived from the depth data.
Trajectories are considered infeasible if user motions exceed the robot's physical speed limits, treating the event of crossing the threshold as a collision.

\begin{figure}[tb!]
\vspace{1.7mm}
  \centering
  \begin{subfigure}[t]{0.48\linewidth}
    \centering
    \includegraphics[width=\linewidth]{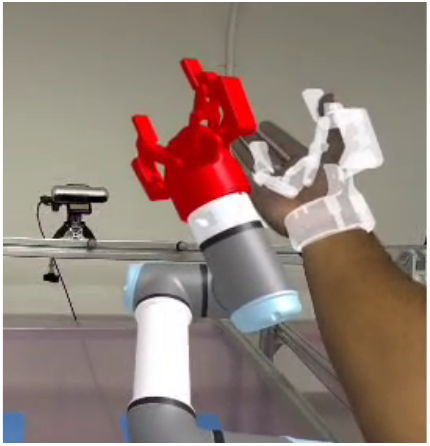}
    \caption{}
  \end{subfigure}%
  \hfill
  \begin{subfigure}[t]{0.48\linewidth}
    \centering
    \includegraphics[width=\linewidth]{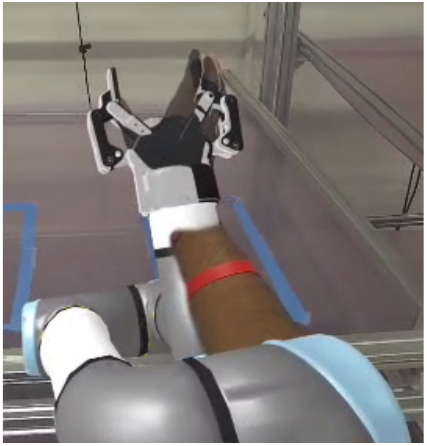}
    \caption{.}
  \end{subfigure}
  \caption{(a) The end effector is colored red due to an infeasible robot pose, waiting for the user to realign. (b) The user's arm is not occluded to prevent human collision with the shelf edge, while the virtual hand is still visible.}
  \label{figure_feasible}
\end{figure}

When a collision or infeasible hand pose occurs, the AR robot pauses tracking, the end effector turns red, and the system reverts to the last valid pose.
Then, the user is prompted to return to that pose—preventing jumps in the recorded trajectory—before either switching to an alternative joint configuration better suited to the current situation or selecting a different motion trajectory. While the system is awaiting alignment, recording is automatically paused. 
Derived depth maps also prevent occlusion of the user’s arm near adjacent obstacles, ensuring safe operation and avoiding human collisions with the environment. The tracked hand remains visualized, allowing the user to assess how closely the virtual pose matches their own. The implemented feedback mechanism (see Figure \ref{figure_feasible}) ensures safety while enabling the collection of consistent, executable demonstrations.

\subsubsection{Demonstration Recording}  
Before starting data collection, ARMimic automatically initializes the coordinate system using gravity for the vertical axis and the forward-facing direction of the headset for the $z$-axis. Since all poses are stored relative to the camera frame, no further calibration is required. A data sampling frequency of 10~\si{\hertz} was found to be sufficient. Due to its intuitive user control and real-time visual feedback, ARMimic enables efficient and unambiguous recording of demonstration episodes.

\subsection{Imitation Learning}

\subsubsection{Data Processing} 
\begin{figure}[tb!]
\vspace{1.7mm}
      \centering
      \includegraphics[width = \linewidth]{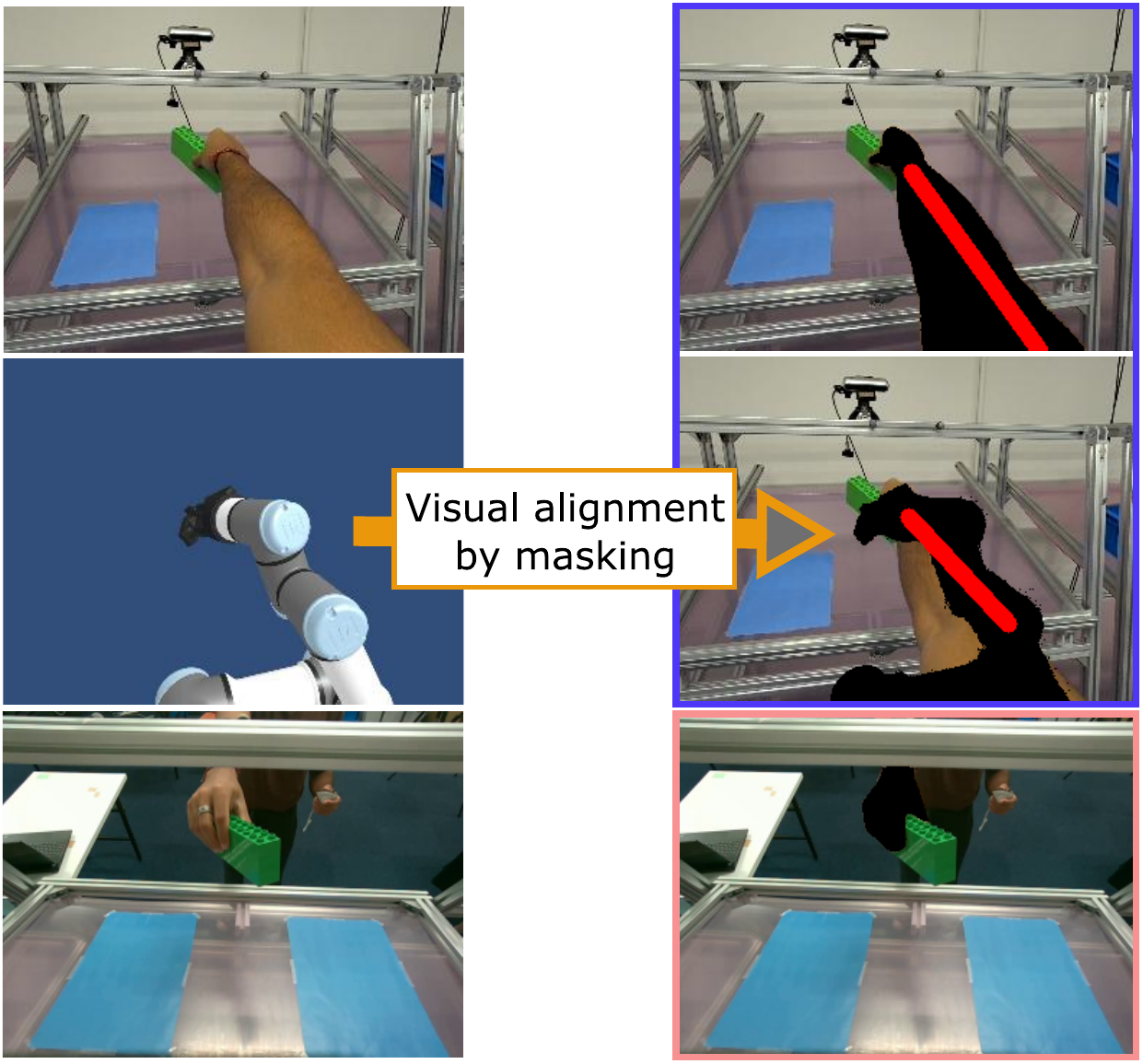} 
      \caption{Visual alignment by masking for the virtual robot and human hand in the headset and external camera views.}
      \label{figure_masking}
\end{figure}

Postprocessing the collected data is necessary for cross-domain alignment, which ensures that both human and robot modalities share a unified representation, allowing the learning algorithm to treat them interchangeably. 
We perform this process by first masking out the robot overlay and human hand in the image frames using the Segment Anything Model 2 (SAM2)~\cite{ravi_sam_2024} and a dedicated hand segmentation model~\cite{camporese2021HandsSeg}, as shown in Figure~\ref{figure_masking}. 
Then, we draw a red line from the wrist to the shoulder for the human and from the end effector to the third-last joint for the robot to indicate the directionality of both arms, which we found to perform best given the robot arm’s non-human morphology.

We temporally align the robot trajectory and corresponding visual contexts from both views and format them into sequences of observation–action pairs~$(a_t,o_t)$.
The policy action at timestep~$t$ corresponds to a sequence of joint angles~${a_{t:t+h}^q=\big(q_t,\dots,q_{t+h})}$ or end effector poses~${a_{t:t+h}^p=\big(p_t,\dots,p_{t+h})}$ of horizon length~$h$. 
To maintain consistent sequence length during training, we apply padding to the actions at the end of a trajectory.

\subsubsection{Model Structure}
\begin{figure}[tb!]
    \vspace{1.7mm}
      \centering
      \includegraphics[width = \linewidth]{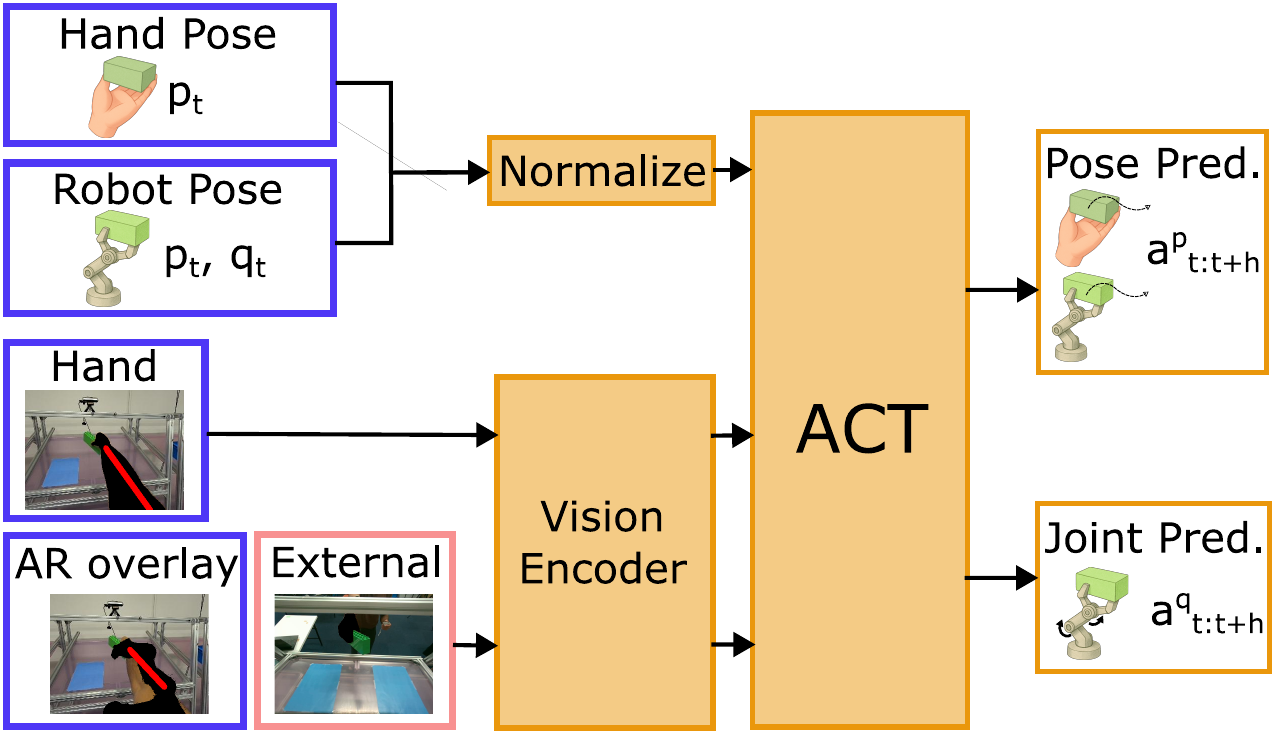} 
      \caption{Model architecture of the joint human-robot policy learning framework of ARMimic. }
      \label{figure_model}
\end{figure}

We can use the processed dataset to train an imitation learning policy (see Figure~\ref{figure_model}).
While different policy formulations are possible, we use Action
Chunking with Transformers (ACT)~\cite{zhao_learning_2023} in this work due to its demonstrated ability to perform complex, long-horizon manipulation tasks.
As observations, we include visual inputs from both external and egocentric views, along with proprioceptive features such as the hand pose and robot joint angles. 
Unlike EgoMimic~\cite{kareer_egomimic:_2024}, we 
normalize human and robot actions using a shared scheme, since their action spaces are identical.
In this way, our approach enables consistent action representation across modalities.
Unlike the original ACT design, where joint prediction is optional, our setup requires it because users directly control the joint configuration. 
Thus, the policy is trained to predict safe and feasible trajectories, reducing the risk of collisions. 
Moreover, we employ an external camera rather than a wrist-mounted camera, providing a more consistent and informative perspective for policy learning.

\subsection{Deployment}
\begin{figure}[tb!]
    \vspace{1.7mm}
      \centering
      \includegraphics[width = 0.8\linewidth]{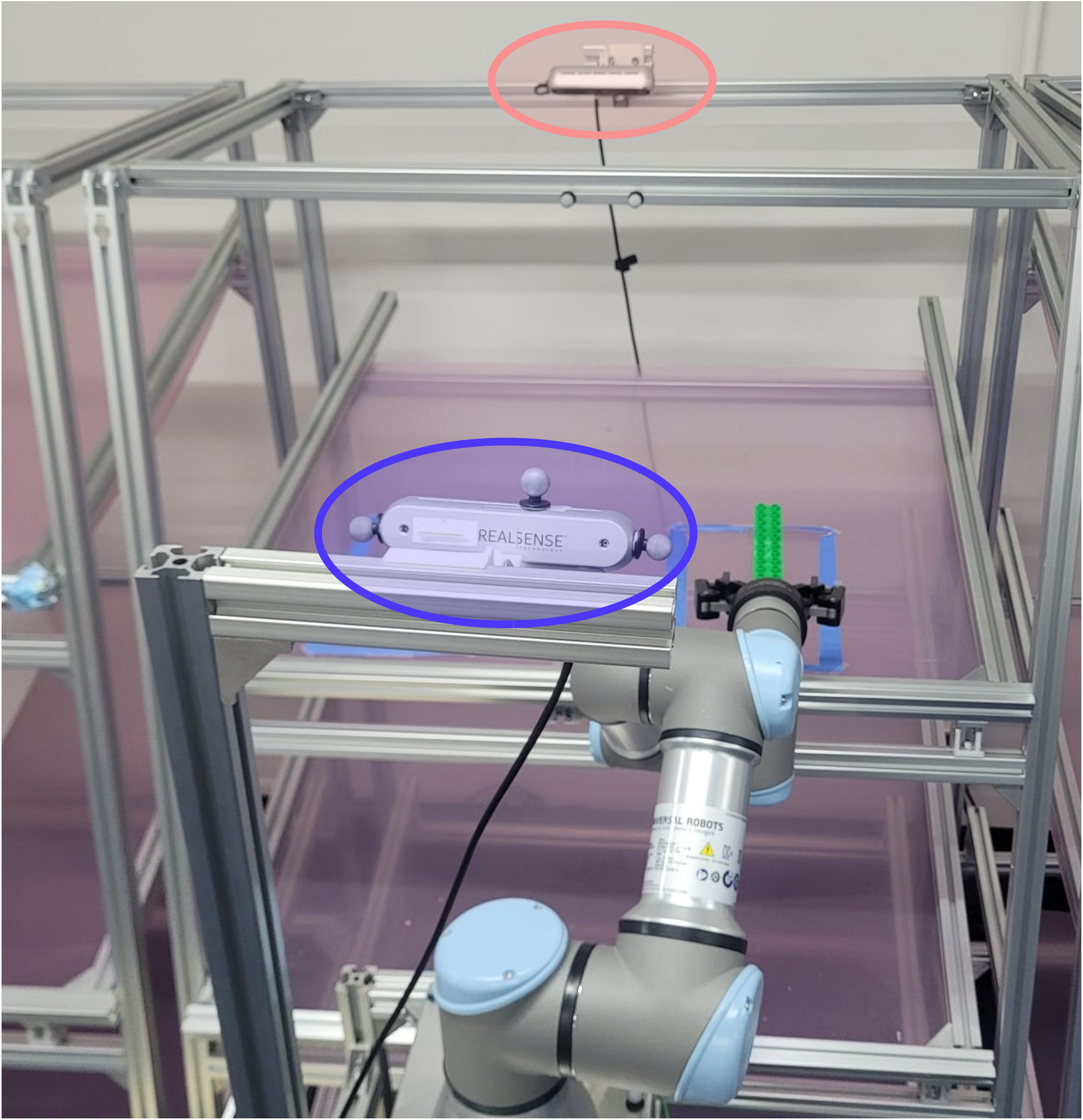} 
      \caption{Two-camera setup to obtain similar egocentric and external view images as during data collection. }
      \label{figure_deploy}
\end{figure}

For deployment on the physical robot, we match the input data streams to those used during dataset collection. The two-view setup—egocentric and external—is replicated by mounting an Intel RealSense D455 RGB-D camera on the robot base, positioned to approximate the XR headset’s viewpoint and ensure visual consistency, as visualized in Figure~\ref{figure_deploy}. The camera frame relative to the robot has to be calibrated only once, for example, using motion capture. All images are preprocessed identically to training before being passed to the policy. Robot state information, including the end effector pose and joint positions, is obtained from onboard sensors.

We deploy ARMimic on a Universal Robot UR3e (Universal Robots A/S, Odense, Denmark)~\cite{ur3e} robot, equipped with a parallel gripper operated in binary open/close mode. 
We set the control frequency of the robot to 25 Hz, with smoothing applied to reduce jitter. From each predicted action chunk of length~$h=100$, only 25 actions are executed before replanning, and we ensure consistent closed-loop behavior by combining current predictions with previous ones via temporal ensembling~\cite{zhao_learning_2023}.

\section{EXPERIMENTS}
\label{sec:EXPERIMENTS}

We conduct extensive real-world experiments, primarily aimed at evaluating the following hypotheses:
\begin{itemize}
\item \textbf{Data collection:} ARMimic enables efficient collection of robust, high-quality datasets.
\item \textbf{Task performance:} Leveraging egocentric hand data with ARMimic improves policy performance on typical manipulation tasks.
\item \textbf{Generalization:} By exploiting human hand data, ARMimic enhances policy generalization to color variations and cluttered scenes, showing performance gains as additional human data is incorporated.
\end{itemize}

\subsection{Experimental Setup}
\subsubsection{Tasks} We consider two challenging real-world manipulation tasks, testing both pick-and-place and stacking capabilities.

\textsc{Pickplace} requires the robot to pick up a tall green toy block from the left (as seen from the workspace camera) region of interest (ROI), which is marked by a blue boundary, and place it into the right ROI. We vary the initial pose of the block, and the robot must ensure that the block is placed securely within the bounds without tipping over.  

\textsc{Stack} is a long-horizon, multi-stage task in which the robot first picks up a styrofoam bowl from the left ROI and places it into the middle bowl located between the two ROIs. It then repeats the process with the bowl from the right ROI. A successful outcome requires all three bowls to be nested within each other, with their bases approximately horizontal. The initial positions of the three bowls may vary within their respective regions.  

Common failure modes in both tasks include incomplete or angled grasps, which hinder precise placement. The \textsc{Stack} task is particularly sensitive to such errors: Even slight misalignments during placement can prevent the bowls from nesting properly, making subsequent stacking attempts difficult or impossible.  

\begin{figure*}[tb!]
    \vspace{1.7mm}
    \centering
    \begin{subfigure}{\textwidth}
        \centering
        \includegraphics[width=\linewidth]{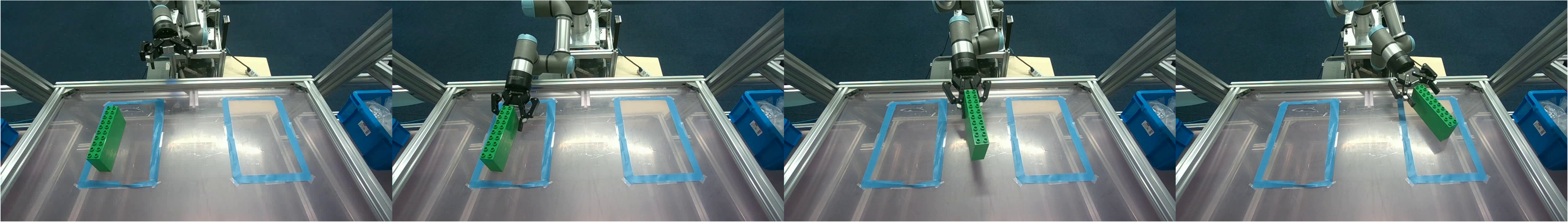}
        \caption{}
        \label{fig:controller_a}
    \end{subfigure}
    
    \vspace{1ex} 
    
    \begin{subfigure}{\linewidth}
        \centering
        \includegraphics[width=\linewidth]{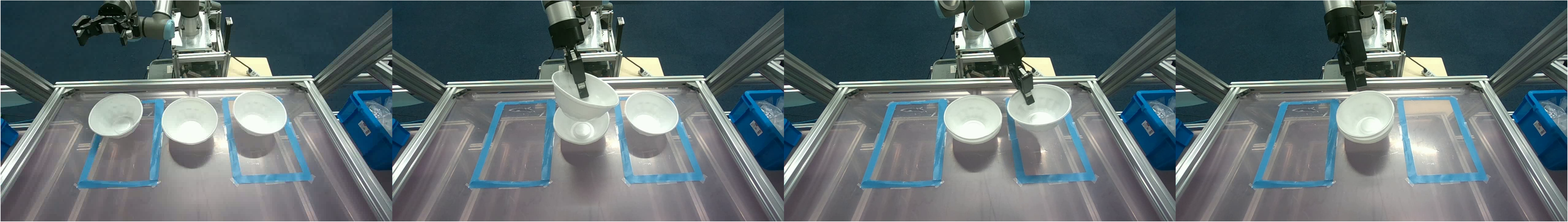}
        \caption{}
        \label{fig:controller_b}
    \end{subfigure}
    
    \caption{Task progression: (a) Pick and place block (b) Stacking three bowls.}
    \label{figure_tasks}
\end{figure*}

Both tasks are carried out on a standard workbench (see Figure \ref{figure_tasks}) under uniform lighting conditions. The external camera remains fixed throughout all trials to ensure visual consistency. To encourage robustness, the initial poses of both the robot base and end effector are varied during data collection.

\subsubsection{Baseline} To contextualize ARMimic's performance, we compare it against an established state-of-the-art baseline, ACT~\cite{zhao_learning_2023}, trained on teleoperation data. For this baseline, the robot is teleoperated using a motion-capture glove that directly maps the human hand's pose to the robot end effector, while the gripper is controlled via keyboard inputs. The input data streams consist of images from the two-camera setup, which ARMimic only uses for deployment, and the robot's proprioceptive measurements.

\subsubsection{Training} 
Both models were trained for 50,000 iterations with a global batch size of 32 on an NVIDIA RTX 4090 GPU. Each training session took approximately 1.5 hours. The hyperparameters used are listed in Table \ref{table:arMimic_training}.

\begin{table}[tb!]
\centering
\small
\caption{Architectural and training details of ARMimic.}
\label{table:arMimic_training}
\begin{tabular}{lc}
\toprule
\textbf{Parameter} & Type/Value \\
\midrule
\textbf{Policy} & ACT~\cite{zhao_learning_2023} \\
\textbf{Batch Size} & 32 \\
\textbf{Optimizer} & AdamW \\
\textbf{Initial Learning Rate} & $5\times10^{-5}$ \\
\textbf{Decay Factor} & 1 \\
\textbf{Scheduler} & Linear \\
\textbf{Encoder Layers} & 4 \\
\textbf{Decoder Layers} & 7 \\
\textbf{Hidden Dim.} & 512 \\
\textbf{Feedforward Dim.} & 3200 \\
\textbf{No. of Heads} & 8 \\
\textbf{Data Augmentations} & Color Jitter \\
\bottomrule
\end{tabular}
\end{table}

\subsection{Results}

\subsubsection{Data Collection} 
We compare collection demonstration datasets using ARMimic against teleoperation and report the results in~Table \ref{table:demo_stats}. 
Collection time is reduced by about 50\% compared to teleoperation, and the replayed trajectories remain collision-free with high success rates. The slightly lower percentage for the \textsc{Stack} task is due to the sensitivity of the gripper alignment: Even small deviations when picking up the bowl can noticeably affect its orientation, leading to task failure.

{\setlength{\tabcolsep}{4.5pt}
\begin{table}[tb!]
\centering
\small
\caption{Data collection with ARMimic compared to teleoperation. Collection time is about~50\% lower with ARMimic, and the recorded trajectories are of high quality, indicated by their high replay success rate~(SR).}
\label{table:demo_stats}
\begin{tabular}{lcc}
\toprule
\textbf{Metric} & \textsc{Pickplace} & \textsc{Stack} \\
\midrule
\textbf{Demo \#}                        & 50   & 50   \\
\textbf{Replay SR$\,$(\%) - ARMimic}              & 96   & 82   \\
\midrule
\textbf{Avg Time (s) - ARMimic}         & \textbf{10.4} & \textbf{15.7} \\
\textbf{Avg Time (s) - Teleoperation}   & 23.1 & 29.1 \\
\bottomrule
\end{tabular}
\end{table}}

\subsubsection{Task Performance}
We evaluate policies trained with ACT and ARMimic over 30 rollouts per seed. 
We use two seeds, with random initial positions for the robot end effector, robot base, and manipulation object within their respective bounds.
As shown by the results in Table~\ref{table:task_method_comparison}, ARMimic achieves an average task success rate of 70\%, which is 11\% higher than ACT.
At the same time, average completion time is reduced by 32\% or, in absolute terms, by approximately~18~\si{\second}.

{\setlength{\tabcolsep}{5.5pt}
\begin{table}[tb!]
\centering
\small
\caption{Performance comparison of ARMimic and ACT, which is trained using teleoperated data.
ARMimic achieves 11\% higher task success and 32\% faster task completion.
}
\label{table:task_method_comparison}
\begin{tabular}{l l c c}
\toprule
\textbf{Task} & \textbf{Method} & \textbf{Success Rate (\%)} & \textbf{Avg Time (s)} \\
\midrule
\multirow{2}{*}{\textsc{Pickplace}} & ACT     & 65 & 47.7 \\
                            & ARMimic & \textbf{75} & \textbf{30.8} \\
\midrule
\multirow{2}{*}{\textsc{Stack}}     & ACT     & 42 & 68.5 \\
                            & ARMimic & \textbf{53} & \textbf{49.6} \\
\bottomrule
\end{tabular}
\end{table}}

\subsubsection{Data Scaling} 
\begin{figure}[t!]
    \vspace{1.7mm}
      \centering
      \includegraphics[width = 0.94\linewidth]{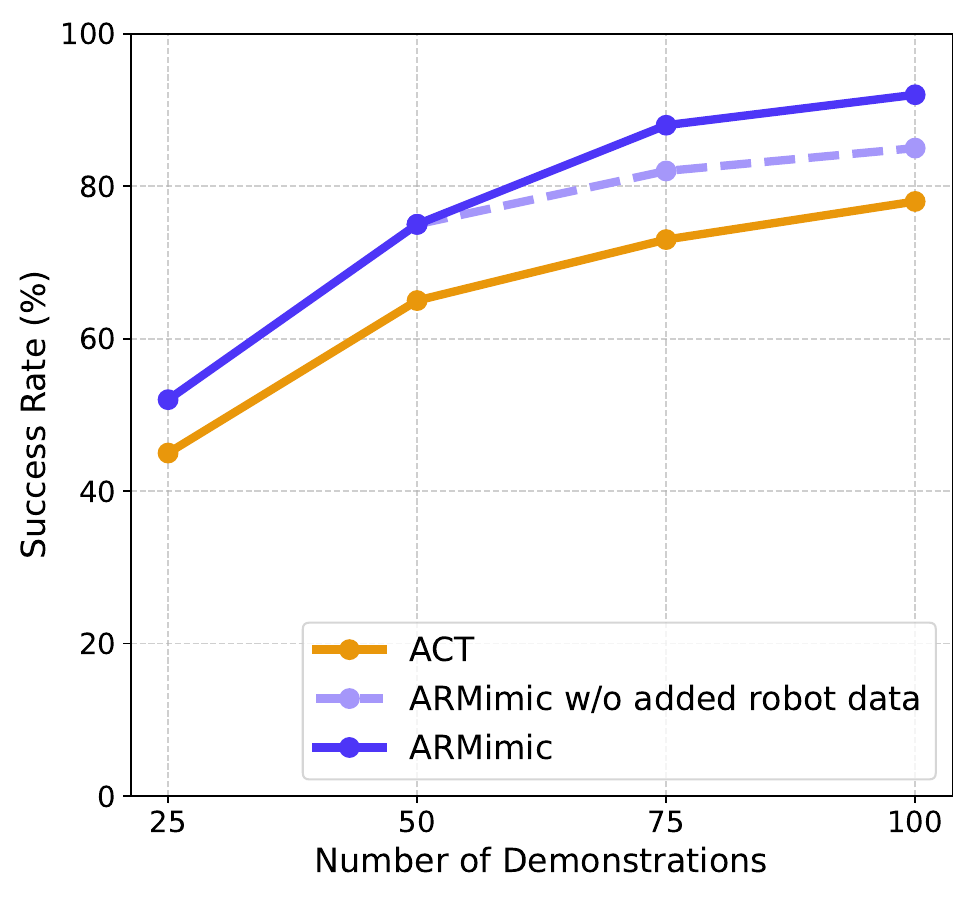} 
      \caption{Impact of scaling the dataset size on task success. ARMimic outperforms ACT by a large margin, which increases as more data is collected.}
      \label{figure_scaling}
\end{figure}

We also evaluate how the performance of ARMimic scales with increasing dataset size and provide the results in~Figure~\ref{figure_scaling}.
Our method consistently surpasses ACT, with the performance gap widening as more demonstrations are added. Interestingly, even when omitting AR robot overlay data, i.e., masked image and robot joints, and adding only 50 human hand demonstrations, ARMimic achieves a 10\% higher success rate compared to the previous dataset size of 50. When we include both 50 human and 50 robot demonstrations, the absolute performance improvement further increases to 17\%. During rollouts of the model trained without additional robot data, we observe that the robot approaches the block from a lower angle—more characteristic of the human demonstrations—which occasionally causes contact with the workplace floor. This further highlights the importance of ARMimic's collision-avoidance safety measures that provide feedback on infeasible robot poses during data collection.

\subsubsection{Generalization} 
\begin{figure}[t!]
    \vspace{1.7mm}
    \centering
    \begin{subfigure}{\linewidth}
        \centering
        \includegraphics[width=\linewidth]{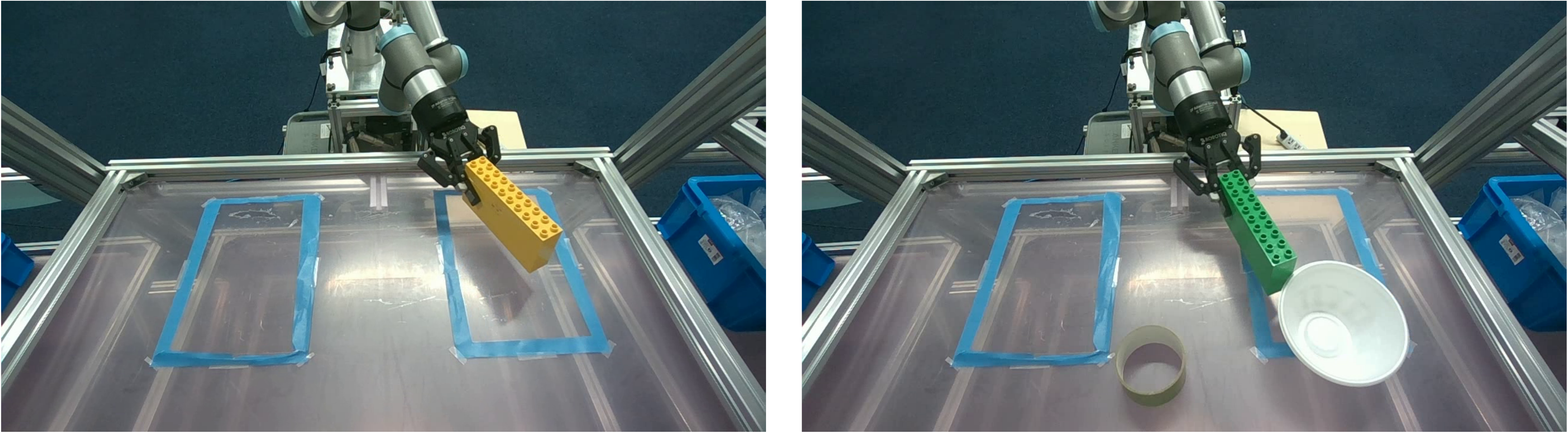}
        \caption{}
        \label{fig:controller_a}
    \end{subfigure}
    
    \vspace{1ex} 
    
    \begin{subfigure}{\linewidth}
        \vspace{1.7mm}
        \centering
        \includegraphics[width=0.95\linewidth]{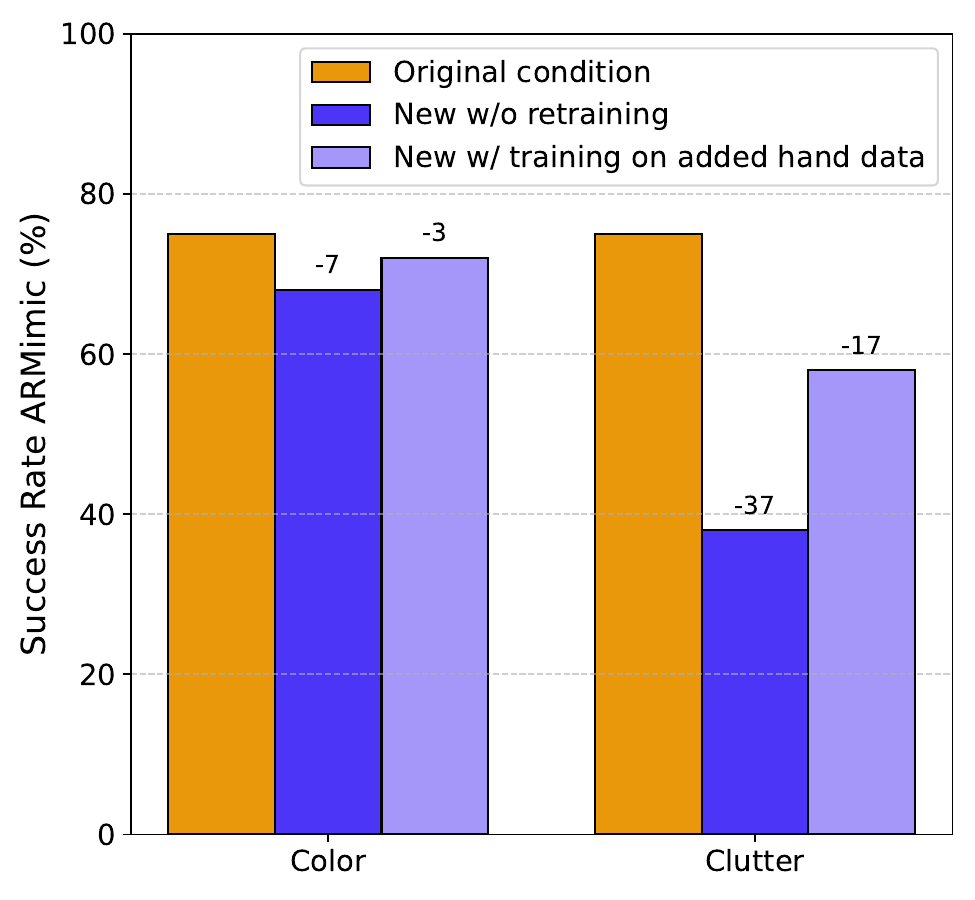}
        \caption{}
        \label{fig:controller_b}
    \end{subfigure}
    
    \caption{(a) Variations in object colors and clutter present in the workspace to evaluate generalization (b) Performance of ARMimic for these different scene conditions with and without retraining on additional hand data collected.}
    \label{figure_general}
\end{figure}

Building on the positive scaling trend of adding human demonstration data without robot data, we investigate whether this strategy can also improve policy generalization.
For this purpose, we introduce object color variations and clutter to the scene, as depicted in Figure~\ref{figure_general}. 
We find that incorporating additional hand data collected under new scene conditions mitigates the drop in success rate by about~50\% in relative terms, particularly in highly cluttered scenes. 
We observe that the policy can better push obstructing objects aside or avoid them altogether, since this behavior is demonstrated in the added hand data.

\section{DISCUSSION}

Experiments on the \textsc{Pickplace} and \textsc{Stack} tasks demonstrate that ARMimic significantly reduces data collection time by about~50\% compared to teleoperation, while maintaining collision-free trajectories. This efficiency stems from the egocentric data collection setup, which removes the sensory gap of teleoperation and makes interactions more intuitive. In combination with extensive feasibility safeguards, this results in smoother and safer demonstrations.

Policies learned using ARMimic outperform the state-of-the-art baseline ACT trained on teleoperation data, achieving higher success rates and much faster task completion. Performance consistently improves as the dataset size increases, with the gap widening relative to ACT. 
Notably, even hand-only demonstrations (without AR robot overlays) improve performance and help policies generalize to new object colors and cluttered scenes.
This robot-agnostic property allows us to first collect a smaller dataset with both AR robot overlay and hand data to establish the joint human-robot representation, and then expand using only human hand demonstrations to improve performance or generalize to new scenes and objects. This is beneficial because collecting the additional hand-only data is simpler and can be carried out by novices, while remaining applicable across different robots for which a human-robot dataset exists.

\section{CONCLUSIONS AND FUTURE WORK}

This work introduces ARMimic, a novel framework for egocentric, hardware-minimalistic data collection using an XR headset and a workspace camera. By combining native hand tracking on the Meta Quest~3 with a virtual robot overlay, ARMimic enables efficient and aligned capture of both human hand motion and robot data—without requiring physical robots or external motion-capture systems. Key features of the system—including real-time collision handling, virtual embodiment correction, and seamless data alignment—ensure high-quality, robust demonstrations while minimizing disruption to the user’s workflow.
In summary, ARMimic is a practical framework for imitation learning, enabling high-quality dataset collection with minimal hardware while supporting scalable and generalizable robot skill acquisition.

There are multiple avenues for future work, such as evaluating ARMimic on even longer-horizon tasks and testing its effectiveness across different robot embodiments with varying morphologies (e.g., humanoids) to investigate the extent to which alignment between human and robot is possible. 
Future user studies could evaluate how novices interact with the AR overlay and how effectively they follow system guidance to produce high-quality demonstrations.
Insights from these studies could support the adoption of ARMimic for scalable, robot-agnostic data collection, accelerating the deployment of autonomous systems in diverse real-world environments.

\newpage
\bibliography{Settings/mybibfile}

\

\addtolength{\textheight}{-12cm}   



\end{document}